\documentclass[letterpaper, 10 pt, conference]{ieeeconf}  

\IEEEoverridecommandlockouts                              

\usepackage{cite}
\usepackage{amsmath,amssymb,amsfonts}
\usepackage{graphicx}
\usepackage{textcomp}
\usepackage{xcolor}
\usepackage{multirow}
\usepackage{multicol}
\usepackage{array}
\usepackage{makecell}
\usepackage{hhline}
\usepackage{mathtools}
\usepackage{booktabs} 
\usepackage{url}
\usepackage{tikz}
\usepackage{algorithm,algpseudocode}
\usepackage[labelformat=empty]{subfig}

\let\labelindent\relax
\usepackage[inline,shortlabels]{enumitem}

\begin{document}

\title{
	CTS: Sim-to-Real Unsupervised Domain Adaptation on 3D Detection
	\thanks{Meiying Zhang and Weiyuan Peng are co-first authors; Corresponding author: Qi Hao (e-mail: hao.q@sustech.edu.cn)}
	\thanks{$^1$ Research Institute of Trustworthy Autonomous Systems, Southern University of Science and Technology (SUSTech), China; $^2$ Department of Computer Science and Engineering, SUSTech; $^3$ Kuang-Chi Institute of Advanced Technology, China.}
	\thanks{This work is jointly supported by the National Natural Science Foundation of China (62261160654), the Shenzhen Fundamental Research Program (JCYJ20220818103006012, KJZD20231023092600001), and the Shenzhen Key Laboratory of Robotics and Computer Vision (ZDSYS20220330160557001).}
}
\author{Meiying Zhang$^1$, Weiyuan Peng$^1$, Guangyao Ding$^2$, Chenyang Lei$^2$, Chunlin Ji$^3$, Qi Hao$^{2}$}

\maketitle
\begin{abstract}
	Simulation data can be accurately labeled and have been expected to improve the performance of data-driven algorithms, including object detection.  However, due to the various domain inconsistencies from simulation to reality (sim-to-real), cross-domain object detection algorithms usually suffer from dramatic performance drops.  While numerous unsupervised domain adaptation (UDA) methods have been developed to address cross-domain tasks between real-world datasets, progress in sim-to-real remains limited.  This paper presents a novel Complex-to-Simple (CTS) framework to transfer models from labeled simulation (source) to unlabeled reality (target) domains. Based on a two-stage detector, the novelty of this work is threefold:
	\begin{enumerate*}
		\item developing fixed-size anchor heads and RoI augmentation to address size bias and feature diversity between two domains,  thereby improving the quality of pseudo-label;
		\item developing a novel corner-format representation of aleatoric uncertainty (AU) for the bounding box,  to uniformly quantify pseudo-label quality;
		\item developing a noise-aware mean teacher domain adaptation method based on AU, as well as object-level and frame-level sampling strategies, to migrate the impact of noisy labels.
		      Experimental results demonstrate that our proposed approach significantly enhances the sim-to-real domain adaptation capability of 3D object detection models, outperforming state-of-the-art cross-domain algorithms, which are usually developed for real-to-real UDA tasks.
	\end{enumerate*}
\end{abstract}

\section{Introduction}

Unsupervised domain adaptation (UDA) research in 3D object detection has yielded outstanding results in various real-world datasets \cite{ganinUnsupervisedDomainAdaptation2015, ganinDomainadversarialTrainingNeural2016a,longLearningTransferableFeatures2015a,chenAdversariallearnedLossDomain2020,tarvainenMeanTeachersAre2017,luoUnsupervisedDomainAdaptive2021,yangSt3dSelftrainingUnsupervised2021,yangST3DDenoisedSelftraining2022}.
By contrast, the sim-to-real domain adaptation has not made much progress yet.
This is primarily due to the point cloud generated in commonly used simulation environments, such as CARLA \cite{dosovitskiyCARLAOpenUrban2017}, have limitations including:
\begin{enumerate*}
	\item ideal and densely collected with minimal noise;
	\item significant statistical disparities from real-world data, as simulated assets are limited in types and sizes; and
	\item insufficient diversity in object features.
	      These limits degrade the sim-to-real domain adaptation performance in 3D object detection.
\end{enumerate*}
\begin{figure}[!tbp]
	\centering
	\includegraphics[width=\columnwidth]{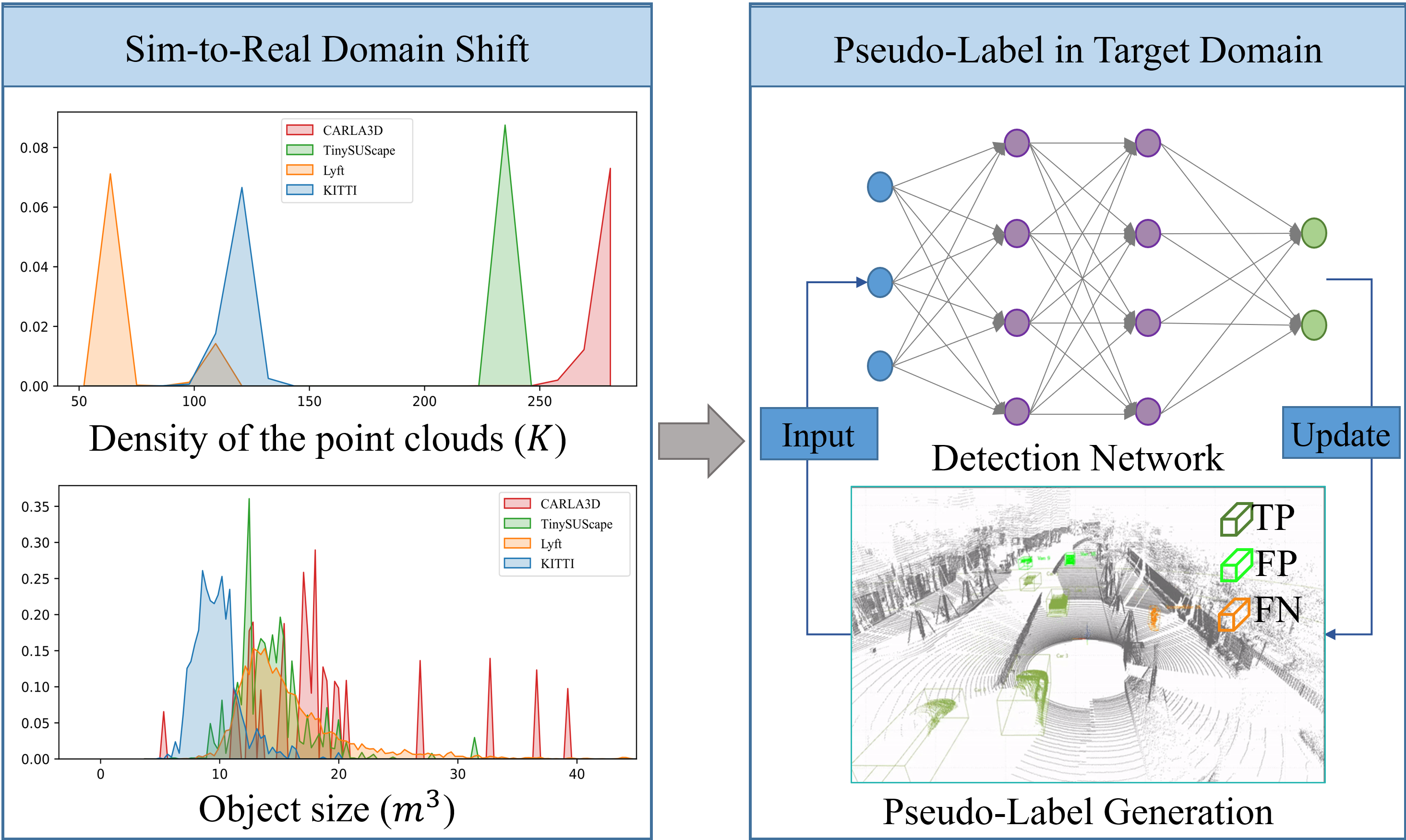}
	\caption{An illustration of unsupervised sim-to-real domain adaptation guided by pseudo-label, which aims to minimize domain shifts arising from the simulator (\textit{e.g.}, CARLA\cite{dosovitskiyCARLAOpenUrban2017}) to the real-world datasets (\textit{e.g.}, KITTI\cite{geigerAreWeReady2012}, Lyft\cite{kestenLyftLevelPerception2019} and TinySUSCape\cite{dingJstJointSelftraining2022}).}
	\label{fig:statistics}
\end{figure}

Generally, UDA methods in 3D object detection can be divided into two main categories:
\begin{enumerate*}
	\item domain-invariant feature learning\cite{ganinUnsupervisedDomainAdaptation2015, ganinDomainadversarialTrainingNeural2016a, longLearningTransferableFeatures2015a, chenAdversariallearnedLossDomain2020}, which learns domain-invariant features by minimizing the distance of feature distribution between the source and target domains;
	\item pseudo-label guided methods\cite{tarvainenMeanTeachersAre2017, luoUnsupervisedDomainAdaptive2021, yangSt3dSelftrainingUnsupervised2021, yangST3DDenoisedSelftraining2022}, which enhance transfer performance by generating pseudo-labels in the target domain and further training using these labels.
\end{enumerate*}
While the former requires specific feature information of two domains, the latter provides a more general and flexible cross-domain framework.
However, these methods are not directly applicable to sim-to-real scenarios. A fully functional pseudo-label guided approach to sim-to-real UDA should be able to address the following issues:
\begin{itemize}
	\item \textbf{Generation of High-quality Pseudo-label.} The object size bias and distribution differences between the simulated and real data, as shown in Fig \ref{fig:statistics}, easily lead to inconsistent regression results (\textit{i.e.}, low-quality pseudo-labels). How to mitigate these biases in detection is important for generating high-quality pseudo-labels.
	\item \textbf{Uniform Quantification of Pseudo-label Quality.}
	      The generated pseudo-labels include true positive (TP), false positive (FP), and false negative (FN), as shown in Fig. \ref{fig:statistics}. In general, TP labels have high quality, while FP ones have low quality and FN ones are missing labels. How to uniformly quantify the quality of pseudo-labels is critical for subsequent sampling of high-quality labels.
	\item \textbf{Target Data Sampling with High-quality Pseudo-labels.} In most UDA methods guided by pseudo-labels, all pseudo-labels are packaged into the target domain training stage. However, FP and FN pseudo-labels introduce extra noise into this process and degrade model performance. How to smartly sample the target data with high-quality pseudo-labels is crucial to improve cross-domain performance.
\end{itemize}

To reduce the domain gap arising from object bias, current methods primarily focus on point cloud preprocessing in the source domain. However, these methods can barely reduce domain inconsistencies between two domains \cite{wangTrainGermanyTest2020,yangST3DDenoisedSelftraining2022,yangSt3dSelftrainingUnsupervised2021}.
Furthermore, methods that use a complex two-stage UDA design show limited performance in sim-to-real tasks \cite{wangTrainGermanyTest2020,luoUnsupervisedDomainAdaptive2021}.
Meanwhile, various methods have been proposed to achieve high-quality pseudo-label guidance, including multi-output fusion techniques, such as fusing multi-modality outputs for 2D-3D data \cite{dingJstJointSelftraining2022}, or fusing multi-pass outputs to maintain ``high stochastic'' \cite{hegde2021uncertainty}.
The mean teacher scheme can also generate more accurate pseudo-labels in target domains \cite{hegde2021uncertainty, luoUnsupervisedDomainAdaptive2021,dengUnbiasedMeanTeacher2021}.
However, its performance can be much degraded by the data noise in sim-to-real tasks.

This paper proposes a mean teacher-based Complex-to-Simple (CTS) framework, focusing on the second stage design, for sim-to-real UDA, with novel techniques to mitigate object bias, enhance pseudo-label quality, and optimize target domain data sampling for pseudo-label guidance.
The main contributions include:
\begin{itemize}
	\item Development of localization refinement techniques including RoI random scaling and fixed-size anchor heads to address domain inconsistencies and produce high-quality pseudo-labels.
	\item Development of a uniform corner-format measure for aleatoric uncertainty (AU) estimation to evaluate the quality of pseudo-labels accurately.
	\item Development of two AU-based sampling strategies in the mean teacher domain adaptation process to select point cloud frames and labels with adequate quality.
	\item Release of CTS code, alongside the CARLA3D simulated dataset, for further research\footnote{The code of CTS and CARALA3D dataset are available at \url{https://github.com/tendo518/CTS-UDA}}.
\end{itemize}

\section{Related Work}\label{sec:related_work}

\subsection{UDA for 3D object detection}

Some previous works have well explored the usage of UDA in 3D object detection \cite{yangSt3dSelftrainingUnsupervised2021,yangST3DDenoisedSelftraining2022, wangTrainGermanyTest2020, luoUnsupervisedDomainAdaptive2021, hegde2021uncertainty}. 
One common challenge of UDA in 3D object detection is the object size bias when cross-domains.
Wang et al.\cite{wangTrainGermanyTest2020} propose statistical normalization (SN) to align object sizes utilizing statistical information from target domain data.
ST3D\cite{yangSt3dSelftrainingUnsupervised2021} and ST3D++\cite{yangST3DDenoisedSelftraining2022} employ data augmentations during source domain training to improve the model's incorporation of diverse size information.
Besides mitigating object size bias, using pseudo-label guided methods in UDA emphasizes improving the quality of pseudo-labels.
JST \cite{dingJstJointSelftraining2022} enhances pseudo-label quality through 2D and 3D joint refinement, aligning outcomes from both modalities.
ST3D \cite{yangSt3dSelftrainingUnsupervised2021} integrates an additional IoU regression head to assess prediction quality, facilitating selective updates of the pseudo-label pool.
Building upon ST3D, ST3D++ \cite{yangST3DDenoisedSelftraining2022} further refines pseudo-labels using a quality-aware denoising pipeline.
MLC-Net\cite{luoUnsupervisedDomainAdaptive2021} also employs the mean teacher scheme to ensure target domain consistency between teacher and student modules at both point and instance levels, which is similar to our method but involves higher complexity using UDA design for both stages.
Although having significant improvements in real-to-real tasks, existing UDA methods often experience serious performance degradation in sim-to-real tasks. Therefore, based on the analysis of simulation and reality differences, our study concentrates on  the quality enhancement, evaluation and selection for pseudo-labels to achieve higher sim-to-real performance.
\begin{figure*}[!t]
	\centering
	\includegraphics[width=0.95\textwidth]{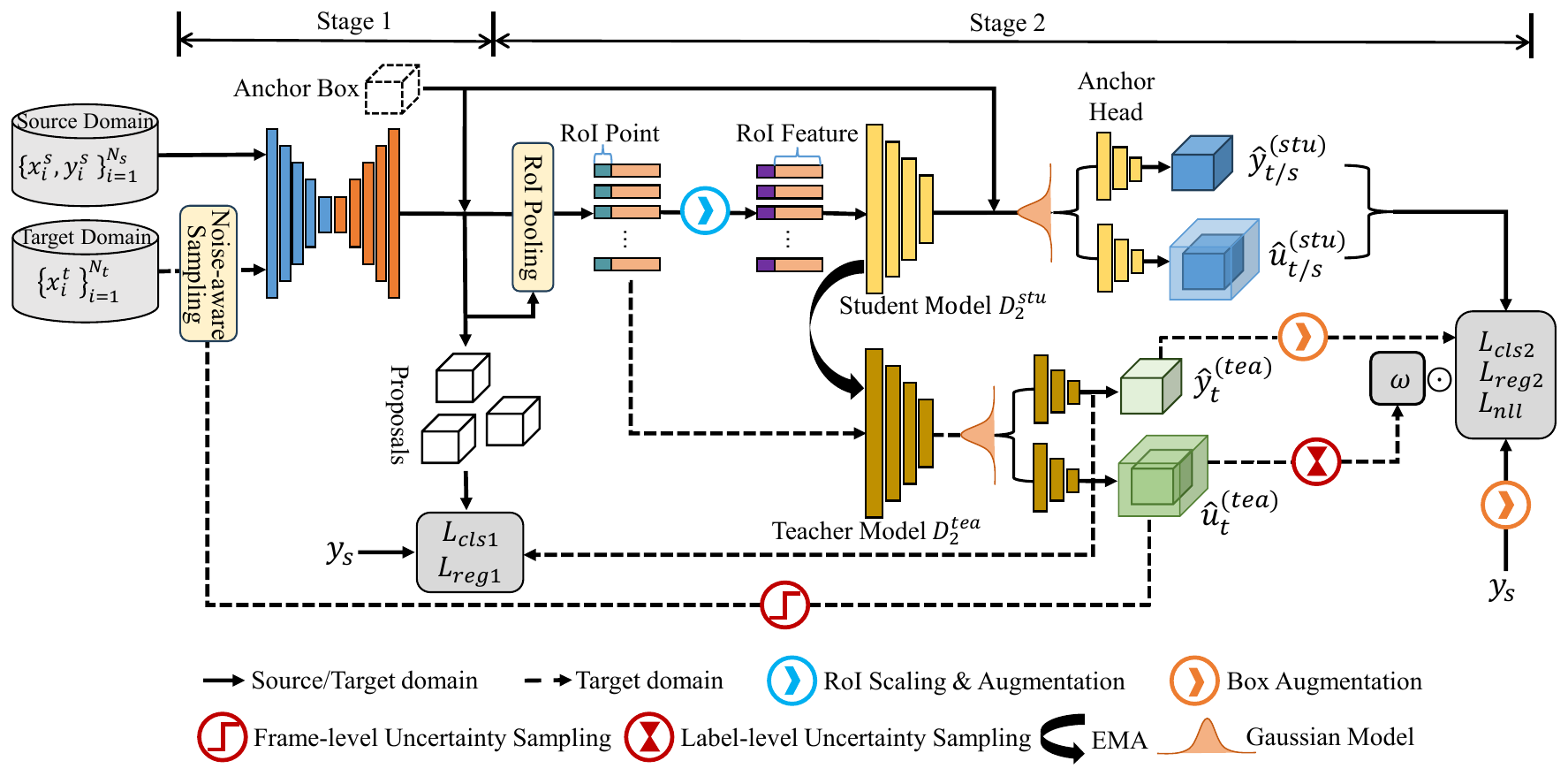}

	\caption{
		An illustration of the CTS framework.
		In the first stage, the model is trained on the source domain with \textit{Anchor Head} (Sec \ref{sec:anchor_head}), \textit{RoI Augmentation} (Sec \ref{sec:roi_augmentation}) and \textit{corner-format AU} modeling (Sec \ref{sec:detection_au}). In the second stage, the \textit{noise-aware} mean teacher approach is applied: the student model is alternatively supervised with pseudo-labels on the target domain and ground-truth labels on the source domain; the teacher model's weights are updated using the EMA. Meanwhile, two \textit{noise-aware} sampling strategies (Sec \ref{sec:noise-aware sampling}) are implemented using the aleatoric uncertainty indicator: frame-level sampling removes noisy frames, while object-level soft-sampling handles noisy labels.
	}
	\label{fig:overview}
\end{figure*}

\subsection{Uncertainty Estimation in 3D Object Detection}

Uncertainty can serve as a valuable metric for quantifying both data and model noise within deep neural networks (DNNs) \cite{kendallWhatUncertaintiesWe2017, galDropoutBayesianApproximation2016, kendallMultitaskLearningUsing2018, tagasovskaSinglemodelUncertaintiesDeep2019, mackayPracticalBayesianFramework1992}. Uncertainty estimation methods typically address two main sources: epistemic uncertainty (EU) and aleatoric uncertainty (AU). EU is represented by a posterior distribution over model parameters\cite{kendallWhatUncertaintiesWe2017, galDropoutBayesianApproximation2016, tagasovskaSinglemodelUncertaintiesDeep2019}, providing insights into the models' uncertainty; AU is represented a distribution over model outputs\cite{kendallMultitaskLearningUsing2018, mackayPracticalBayesianFramework1992}, reflecting intrinsic data stochastic. Notably, AU varies with the quality of input data, suitable for quantifying the noise level of input data.
Within the context of 3D detection tasks, several methodologies have integrated aleatoric uncertainty (AU) due to its ability to enhance detection performance\cite{meyerLasernetEfficientProbabilistic2019, fengLeveragingHeteroscedasticAleatoric2019}. 
Meyer et al.\cite{meyerLasernetEfficientProbabilistic2019} employ a mixture of Laplace distributions to fit the variances for each predefined regression variable, including box center positions, sizes, and orientation. Feng et al.\cite{fengLeveragingHeteroscedasticAleatoric2019} model AU using multivariate Gaussian distributions, with independent variables representing three distinct sets, \textit{i.e.}, RoI positions, bounding box positions, and orientation.
However, few methods have leveraged the Aleatoric Uncertainty (AU) estimated from 3D detection results for the evaluation of data noise. Besides, existing approaches represent uncertainties using non-uniform variables, adding a complexity to further utilization. Therefore, this study proposes a uniform corner-based representation for bounding boxes with uncertainties, easy for the quality evaluation of the predicted pseudo-labels.
\section{System Setup}\label{sec:system_setup}


In a standard two-stage detector like PointRCNN \cite{shiPointrcnn3dObject2019}, the first stage roughly detects objects across a frame and the second stage refines localization. Directly applying PointRCNN for sim-to-real tasks led to a 60\% decrease in Average Precision (AP) at an IoU threshold of 0.7 and a 20\% decrease at IoU of 0.5 (see CARLA3D$\rightarrow$ KITTI in Table \ref{table:main_results}), which suggests a retained object detection and classification ability but much loss in localization precision.
To enhance sim-to-real domain adaptation, the paper simply focuses on improving the domain adaptation of the second-stage localization network instead of adopting a complex two-stage UDA design, namely Complex-to-Simple (CTS).

The complete CTS framework is illustrated in Fig. \ref{fig:overview}. This framework utilizes simulated data from the source domain to initially develop detection capabilities, followed by model refinement through mean teacher-based domain adaptation in real-world traffic scenarios of the target domain. The mean-teacher method \cite{tarvainenMeanTeachersAre2017} mitigates the impact of domain shift by training the student model with a consistency objective, effectively utilizing unlabeled data from the target domain to improve the model's performance in that domain.

The framework consists of two branches: the student and teacher models. Both models share the same architecture and are initialized with parameters trained on the source domain. However, they are updated through different mechanisms:

\subsubsection{Student Model}
The student model utilizes augmented RoI points and features as input, supervised with pseudo-labels $\hat y_t$ in the target domain or ground truth labels $y_s$ in the source domain. It is worth noting that the generated pseudo-labels can serve as supervision for the 1st-stage network as well, thus enabling domain adaptation for the 1st-stage network. Thus, the total loss of this network includes:
\begin{enumerate*}
	\item first-stage RoI regression loss $l_{reg1}$.
	\item first-stage RoI classification loss $l_{cls1}$.
	\item second-stage regression Smooth-L1 loss $l_{reg2}$.
	\item second-stage classification loss $l_{cls2}$.
	\item second-stage AU-NLL loss $l_{nll}$ specified in Sec \ref{sec:detection_au}.
\end{enumerate*}

\subsubsection{Teacher Model}
The teacher model processes raw (non-augmented) data and maintains fixed weights during the backward pass. Instead of employing standard backpropagation with predefined loss functions, the teacher model updates its weights using exponential moving average (EMA) from student model as follows:
\begin{equation}
	\theta^{tea}_{t}=\beta \times \theta^{tea}_{t-1} + (1 - \beta)\times \theta^{stu}_{t}
	\label{eq:EMA}
\end{equation}
Here, $\theta^{stu}_t$ represents the student model's weights at iteration $t$, $\beta$ is the EMA decay factor that controls the update rate, and $\theta^{tea}_t$ denotes the teacher model's weights. The resulting teacher model's weights provide a smoothed representation of the student model's weights over time.

\section{Proposed Methods}\label{sec:proposed_method}

\subsection{Enhancement of Pseudo-Label Quality}
CTS starts by training a detector in a labeled source domain and then leverages this knowledge to generate high-quality pseudo-labels in the target domain. However, simulation-reality differences, such as differences in object size and point density, present significant challenges. 
Specifically, size bias has been shown to significantly reduce localization accuracy\cite{wangTrainGermanyTest2020}, especially when in simulation-to-reality scenarios, where expanding the simulation model asset library (e.g., through CAD modeling) to match the target domain is both difficult and expensive. To address these domain shifts and improve the reliability of pseudo-labels in the target domain, we propose the following methodology:
\subsubsection{Anchor Head (AH)}\label{sec:anchor_head}
The second-stage model typically predicts size residuals $\Delta_{whl}$ between proposals from the first-stage and final bounding boxes, denoted as $\hat{B}$. This approach avoids regressing the size of bounding boxes entirely from scratch. However, a challenge arises when the first-stage model, trained with biased supervision from source domain labels, exhibits inaccuracy in estimating proposal sizes. Unreliable proposal box sizes can lead to size errors accumulating in the second stage, degrading final bounding box refinement accuracy and the effectiveness of pseudo-labels.
Inspired by anchor-based detectors \cite{langPointpillarsFastEncoders2019}, we introduce a fixed-size anchor box ${w_{an}, h_{an}, l_{an}}$ to replace the proposal, termed the \emph{anchor head (AH)}. The AH replaces the traditional proposal mechanism, allowing the second-stage network to work with a globally fixed-size 3D anchor instead of refining proposals. By employing the AH in both the source and target domains, we ensure consistent behavior of the second-stage network regarding proposal size refinement across domains, thus avoid size error propagation in UDA and enhancing the quality of pseudo-labels.

\subsubsection{RoI Random Scaling (RRS) and Augmentation}\label{sec:roi_augmentation}
To enhance the diversity in the features of the learning object from the simulated data, we introduce RoI Random Scaling (RRS) and Augmentation. In our setup, the second-stage model utilizes localized points (RoI points) and corresponding RoI features from the first-stage model as inputs. Specifically, only the points undergo augmentation, while their features remain unchanged. Let $\widetilde{X}\in\mathbb{R}^{3\times N}$ denote the decentralized points within a RoI box of dimensions $l, w, h$, and let $q_{l}, q_{w}, q_{h}$ represent random scaling factors. The scaled RoI sizes are derived by multiplying the original dimensions by the scaling factors, resulting in $q_{l}l, q_{w}w, q_{h}h$.
Furthermore, to enhance the second-stage model's robustness, we apply augmentations that involve random rotation, flipping, and translation within specified ranges, as described in \cite{teamOpenPCDetOpensourceToolbox2020}.

\subsection{3D Detection with Aleatoric Uncertainty}\label{sec:detection_au}
As noted in \cite{kendallWhatUncertaintiesWe2017}, Deep Neural Networks (DNNs) are capable of predicting aleatoric uncertainty effectively. Specifically, in the case where the regression $y$ follows a Gaussian distribution with parameters $(\mu, \sigma^2)$, the following loss function $\mathcal{L}_{nll}$ can be employed for optimization:
\begin{equation}
	\mathcal{L}_{nll}=\frac{(y-f_{\mu}(\mathbf{x}, \theta))^2}{2f_{\sigma^2}(\mathbf{x}, \theta)}+ \frac {1}{2} \log (f_{\sigma^2}(\mathbf{x}, \theta))
\end{equation}
where $\theta$ is the model parameter, $f_{\mu}$ and $f_{\sigma^2}$ represent subnetworks for predicting the mean and the variance.

When training the regression part of the detector, since the predicted bounding box $y$ is usually encoded with 7 values, i.e., $\mathbf{y}_b=\{\mu_{bx}, \mu_{by},\mu_{bz},\mu_{bh},\mu_{bw},\mu_{bl},\mu_{b\alpha}\}$ (called box format, BF), the matched variance values are encoded primarily as $\boldsymbol{\sigma}^2_b=\{\sigma_{bx}^2, \sigma_{by}^2, \sigma_{bz}^2, \sigma_{bh}^2, \sigma_{bw}^2, \sigma_{bl}^2, \sigma_{b\alpha}^2\}$, of which each element corresponding to the uncertainty of an element in the bounding box representation.
Nevertheless, the BF bounding box regression variable, specifically the centroid positions, extents (length, width, height), and orientations exhibits numerical magnitude inconsistencies. These disparities also indicate varying magnitudes of variances across each variable. Applying reduction methods (such as maximum or average) to these variances naively may result in overlooking uncertainties arising from specific components, particularly the orientation, due to its significantly smaller magnitudes.

\begin{figure}
	\centering
	\vspace{-9pt}  
	\includegraphics{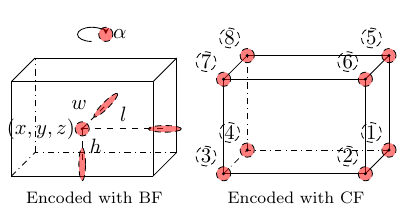}
	\caption{An illustration of two coding schemes of bounding boxes with uncertainties. (a) BF: box format; (b) CF: corner format, where the red areas stand for the potential ranges,  that is, the aleatoric uncertainty.} \label{fig:corner_trans}
\end{figure}

Inspired by the corner loss methodology\cite{chenMultiview3dObject2017}, we introduce a corner-based uncertainty measurement by encoding the bounding box equally with its 8 corner points, as illustrated in Fig \ref{fig:corner_trans}.
To be specific, during the training process, we firstly perform corner transformation on both model-predicated BF box and corresponding ground truth:
\begin{equation}\label{eq:corner_transformation}
	\begin{bmatrix}
		\mu_{cx}^i \\  \mu_{cy}^i \\ \mu_{cz}^i
	\end{bmatrix}
	=
	\textit{R}_{z}(\mu_{b\alpha})
	\times
	\begin{bmatrix}
		\pm\frac{\mu_{bw}}{2} \\  \pm\frac{\mu_{bh}}{2} \\   \pm\frac{\mu_{bl}}{2}
	\end{bmatrix}
	+
	\begin{bmatrix}
		\mu_{bx} \\ \mu_{by} \\ \mu_{bz}
	\end{bmatrix}
\end{equation}
Where ${\textit{R}_{z}(\mu_{b\alpha})}$ represents the rotation matrix corresponding to the yaw angle $\mu_{b\alpha}$, and $p_c^i={\mu_{cx}^i, \mu_{cy}^i, \mu_{cz}^i}_{i=1}^8$ denotes the positions of the 8 corners of the transformed CF-encoded box.
For the sake of regression simplification, we assume that the distribution of each corner's coordinates follows distinct Gaussian all sharing the same variance, denoted as:
\begin{equation}
	\mathbf{y}_c^i=
	\begin{bmatrix}
		y_{cx}^i \\  y_{cy}^i \\ y_{cz}^i
	\end{bmatrix}
	\sim
	\mathcal{N}
	\left(
	\begin{bmatrix}
		\mu_{cx}^i \\ \mu_{cy}^i \\ \mu_{cz}^i
	\end{bmatrix},
	(\sigma_c^i)^2\mathbf{I}
	\right)
	,i=1\ldots8
\end{equation}
where $\mathbf{I}$ is the identity matrix.
Consequently, we predict 8 (rather than 24) independent variances $(\sigma_c^2)^i$ for a CF encoded box, the overall NLL loss $\mathcal{L}$ and aleatoric uncertainty $\hat{u}$ can be easily reduced with:
\begin{align}
	\mathcal{L}_{nll}^i=
	\frac{\left(\overline{\mathbf{y}_c^i}-\overline{\hat{\mathbf{y}}_c^i}\right)^2}
	{2(\sigma_c^i)^2}
	+ \frac {1}{2} \log(\sigma_c^i)^2
\end{align}

Where final NLL loss $\mathcal{L}_{nll} = \frac{\sum_{i=1}^{8}\mathcal{L}_{nll}^i}{8} $ and uncertainty of box $u_{box} =\frac{\sum_{i=1}^{8}(\sigma_c^i)^2}{8}$. And all components contribute equally to the loss and final uncertainty metric.

\subsection{Noise-aware Mean Teacher}\label{sec:noise-aware sampling}

Aligning transformations on both student-model inputs and teacher-model output facilitates the acquisition of domain-invariant representations, thereby aiding in adaptation to the target domain using pseudo-labels. However, noisy pseudo-labels can lead to error accumulation. To address this challenge, we leverage aleatoric uncertainties predicted by a model to annotate data in the target domain and mitigate the impact of noisy data during mean teacher domain adaptation with the following sampling strategies:

\subsubsection{Object-Level Soft Sampling}
During each iteration, the final second-stage regression loss $L_{reg2}$ is computed using the supervision provided by pseudo-labels assigned to individual objects. Rather than solely depending on these pseudo-labels, the loss is weighted by the inverse of their uncertainty $u$, denoted as:
\begin{align}
	 & \mathbf{w}_{label} = \{\frac{1}{u} \mid \forall u \in \hat{\mathbf{u}}_{tea} \}
	 & l_{2} =\mathbf{w}_{label} \odot \mathbf{l}_{2}
\end{align}
Where $\mathbf{l}_{2}$ is the second-stage loss produced per object in the whole point cloud frame, $\odot$ is the element-wise product.
Consequently, objects with higher uncertainty associated with their pseudo-labels are softly filtered out, mitigating the adverse effects of noisy objects.

\subsubsection{Frame-Level Sampling}
Instead of using all target data, the sampling process selects a subset based on frame-level uncertainty. Low-noise target frames are sampled to train the model, enhancing its ability to detect objects in the target domain. By integrating curriculum learning strategies\cite{wang_survey_2021}, the model refines its pseudo-labels and becomes more confident in uncertainty estimates after several training epochs. This iterative process gradually includes more frames until eventually, all target data are sampled. A detailed explanation of the frame-level sampling refers to Algorithm \ref{alg:frame_level}.

\begin{algorithm}[h]
	\caption{Noise-aware Frame-Level Sampling}\label{alg:frame_level}

	\renewcommand{\algorithmicrequire}{\textbf{Input:}}
	\renewcommand{\algorithmicensure}{\textbf{Output:}}
	\begin{algorithmic}[1]
		\Require
		\Statex $\mathcal{T}$: Unlabeled Target Domain Dataset
		\Statex $\mathcal{T}_{sub}$: Target Sub-dataset after Sampling
		\Statex $N_t$: Number of samples in $\mathcal{T}$
		\Statex $N_{sub}$: Amount of data to be selected

		\Ensure $\mathcal{D}$: Noise-aware Model

		\While{$N_{sub} < N_t$}
		\State ${U}_{frame} \gets \{\}$, $\mathcal{T}_{sub} \gets \{\}$
		\For{each frame $x^{t}$ in $\mathcal{T}$}
		\State $\hat{\mathbf{y}}^t, \hat{\mathbf{u}}^t \gets $ inference $\mathcal{D}$ for $x^{t}$
		\State $\hat{u}^t \gets \text{mean of } \hat{\mathbf{u}}^t \text{ for all valid object in $x^{t}$}$
		\State ${U}_{frame} \gets $ append $\hat{u}_t$ to ${U}_{frame} $
		\EndFor

		\For{$i$ in $\{1,\ldots,{N}_{sub}\}$}

		\State $j \gets \text{argmin of } {U}_{frame}$
		\State $\mathcal{T}_{sub} \gets$ append the $j_{th}$ element $x_j^t$ in $\mathcal{T}$ to $\mathcal{T}_{sub}$
		\State pop the $j_{th}$ element $\hat{u}_j^t$ from ${U}_{frame}$
		\EndFor
		\State $\mathcal{D} \gets $ fine-tune $ \mathcal{D}$ with $\mathcal{T}_{sub}$
		\State ${N}_{sub}+={N}_{sub}$
		\EndWhile
		\State \bf{return} $\mathcal{D}$
	\end{algorithmic}

\end{algorithm}

\section{Experiments}\label{sec:experiments}

\begin{table*}[!ht]
	\vspace{0.2cm}
	\renewcommand\arraystretch{1.2}
	\centering
	\begin{tabular}{c|c|ccc|ccc}
		\Xhline{0.9pt}
		\multirow{2.3}{*}{Task} & \multirow{2.3}{*}{Method}                       & \multicolumn{3}{c|}{$AP_{BEV}@0.7$} & \multicolumn{3}{c}{$AP_{3D}@0.7$}                                                                     \\
		\cline{3-8}
		                        &                                                 & Easy                                & Moderate                          & Hard           & Easy           & Moderate       & Hard           \\

		\Xhline{0.5pt}

		\multirow{5}{*}{CARLA3D$\rightarrow$Lyft}
		                        & Source Only                                     & 66.70                               & 54.35                             & 51.76          & 18.82          & 13.85          & 13.64          \\
		                        & SN\cite{wangTrainGermanyTest2020}               & 66.92                               & 53.31                             & 50.52          & 23.05          & 16.79          & 15.99          \\
		                        & MLC-Net\cite{luoUnsupervisedDomainAdaptive2021} & 77.95                               & 64.46                             & 62.13          & 53.97          & 40.04          & 37.47          \\
		                        & ST3D++\cite{yangST3DDenoisedSelftraining2022}   & 75.57                               & 61.68                             & 57.49          & 51.02          & 37.24          & 35.41          \\
		                        & \textbf{Ours}                                   & \textbf{81.66}                      & \textbf{67.86}                    & \textbf{65.17} & \textbf{61.93} & \textbf{45.87} & \textbf{43.87} \\
		\cline{2-8}
		                        & Oracle                                          & 90.92                               & 83.97                             & 81.70          & 80.06          & 66.05          & 64.01          \\
		\Xhline{0.7pt}
		\specialrule{0em}{1pt}{1pt}
		\Xhline{0.7pt}
		\multirow{5}{*}{CARLA3D$\rightarrow$KITTI}
		                        & Source Only                                     & 27.45                               & 20.55                             & 17.51          & 5.67           & 4.06           & 3.23           \\
		                        & SN\cite{wangTrainGermanyTest2020}               & 31.21                               & 30.23                             & 28.18          & 9.37           & 9.15           & 7.63           \\
		                        & MLC-Net\cite{luoUnsupervisedDomainAdaptive2021} & 70.45                               & 56.66                             & 49.41          & 43.02          & 32.68          & 27.39          \\
		                        & ST3D++\cite{yangST3DDenoisedSelftraining2022}   & 64.50                               & 54.91                             & 49.75          & 34.34          & 27.22          & 23.99          \\
		                        & \textbf{Ours}                                   & \textbf{78.92}                      & \textbf{64.17}                    & \textbf{57.37} & \textbf{58.41} & \textbf{45.28} & \textbf{39.61} \\

		\cline{2-8}
		                        & Oracle                                          & 93.18                               & 83.26                             & 80.20          & 86.02          & 71.70          & 66.86          \\
		\Xhline{0.7pt}
		\specialrule{0em}{1pt}{1pt}
		\Xhline{0.7pt}
		\multirow{4}{*}{CARLA3D$\rightarrow$TinySUScape}
		                        & Source Only                                     & 18.02                               & 16.69                             & N/A            & 4.59           & 3.83           & N/A            \\
		                        & SN\cite{wangTrainGermanyTest2020}               & 27.45                               & 14.96                             & N/A            & 1.42           & 1.36           & N/A            \\
		                        & MLC-Net\cite{luoUnsupervisedDomainAdaptive2021} & 19.64                               & 18.81                             & N/A            & 8.27           & 7.59           & N/A            \\
		                        & ST3D++\cite{yangST3DDenoisedSelftraining2022}   & 40.86                               & 38.17                             & N/A            & 26.09          & 23.86          & N/A            \\
		                        & \textbf{Ours}                                   & \textbf{42.45}                      & \textbf{38.62}                    & N/A            & \textbf{31.47} & \textbf{28.02} & N/A            \\
		\Xhline{0.7pt}
	\end{tabular}
	\caption{Comparison results of three different sim-to-real domain adaptation tasks. We report $AP_{BEV}$ and $AP_{3D}$ of the \textit{car} category at IoU = 0.7 for different difficulty levels. As TinySUSCape\cite{dingJstJointSelftraining2022} does not provide labels with the occlusion level, \textit{Hard} is marked as Not Available (N/A).}
	\label{table:main_results}
\end{table*}

\subsection{Experimental Setup}
\subsubsection{Datasets} \label{carla3d_detail}
Most existing LiDAR simulation datasets are primarily used for task-specific problems, such as Vehicle-to-Vehicle Communication \cite{xu2022opv2v} and Continuous Domain Shift \cite{sun2022shift}, rather than for sim-to-real UDA as addressed in this paper.
Thus, we conduct supervised training in a simulated source domain, namely CARLA3D, acquired within the CARLA simulator \cite{dosovitskiyCARLAOpenUrban2017} from scratch.
All samples are taken from eight built-in scenarios in CARLA to ensure data diversity.
The ego-vehicle is positioned randomly, collecting about 100 samples per scenario, each comprising eight frames at 2Hz.
Out of the eight frames per sample, five are randomly chosen for the training set, yielding 3,990 frames with a total of 25,192 objects.
Further details of the CARLA3D dataset are outlined in Table \ref{tab:CARLA3D_dataset}.
The target domains chosen include KITTI \cite{geigerAreWeReady2012}, Lyft \cite{kestenLyftLevelPerception2019}, and TinySUScape used in \cite{dingJstJointSelftraining2022}.
During the testing phase, samples from these datasets along with their corresponding labels will be utilized, whereas only samples will be used during the training phase. A summary of these datasets is presented in Table \ref{tab:datasets}.

\subsubsection{Evaluation Metric}
In our 3D object detection evaluation, referring to \cite{wangTrainGermanyTest2020}, we utilize the official KITTI evaluation metric from \cite{geigerAreWeReady2012} for the \textit{Car} category. We report two average precision (AP) metrics: $AP_{BEV}$ based on bird's-eye view IoUs, and $AP_{3D}$ based on 3D IoUs.

\subsubsection{Implementation Details}
Our proposed method is implemented based on OpenPCDet\cite{teamOpenPCDetOpensourceToolbox2020}, using PointRCNN\cite{shiPointrcnn3dObject2019} as our baseline detector.
All experiments were conducted on a Ubuntu Linux server equipped with 12 GiB NVIDIA TITAN V GPUs.
The proposed model is first trained in CARLA3D for 50 epochs, in which the learning rate, the weight decay, and the momentum are set as 0.005, 0.0001, and 0.9, respectively.
For the anchor head configuration, the anchor dimensions are globally set to $l_{an}=3.9$, $h_{an}=1.6$, and $w_{an}=1.56$. These values are derived from the statistical average of the dimensions of all labeled car objects in the KITTI dataset, deemed a reasonable metric.
RoI augmentation is applied, involving random scaled by a factor of range from $0.7$ to $1.3$, translated by up to $\pm0.5$ meter, rotated by an angle between $-\frac{\pi}{4}$ and $\frac{\pi}{4}$, and flipped by a chance of 50\%.
During mean teacher domain adaptation, the model achieving the highest accuracy in the source domain training phase is selected, and both teacher and student models are initialized from it. The Exponential Moving Average (EMA) factor ($\beta$) is set to 0.999, and the training lasts for 30 epochs for the Lyft dataset and 50 epochs for the KITTI/TinySUScape datasets.
To ensure stability, we train the student model by alternating between source (with ground-truth labels) and target (with pseudo-labels) domain data.
Regarding noise-aware training settings, the uncertainty pool is refreshed at the 1st, 6th, 16th, and 21st epochs for the Lyft dataset and at the 1st, 11th, 21st, and 31st epochs for the KITTI and TinySUScape datasets. In each of these epochs, sub-datasets are resampled at percentages of 30\%, 50\%, 70\%, and 100\% of the total dataset size for subsequent training iterations.

\subsection{Main Results}\label{exp_main_results}

Our CTS framework was compared with the following methods:
\begin{enumerate*}
	\item \textbf{SN}\cite{wangTrainGermanyTest2020}: A domain adaptation method has been considered effective on various datasets;
	\item \textbf{MLC-Net}\cite{luoUnsupervisedDomainAdaptive2021}: A domain adaptation method also based on mean teacher, which is similar to ours in the mean teacher part;
	\item \textbf{ST3D++}\cite{yangST3DDenoisedSelftraining2022}: A recent self-training based method that achieved state-of-the-art performance in real-to-real (\textit{e.g.}, Nuscenes\cite{caesarNuscenesMultimodalDataset2020} $\to$ KITTI\cite{geigerAreWeReady2012}) domain adaptation tasks.
\end{enumerate*}

Besides, we provide two possible boundaries of results, they are:
\begin{enumerate*}
	\item \textbf{Source Only}: The model is solely trained in a supervised manner on the source domain and is directly applied to the target domain without employing any domain adaptation methods, which serve as a lower bound;
	\item \textbf{Oracle}: A fully supervised model trained on the target/reality domain with actual labels, considered as an upper bound.
\end{enumerate*}

\begin{table}[tbp]
	\centering
	\begin{tabular}{cccccc}
		\toprule
		Scenario & Frames & Easy   & Moderate & Hard    & Times \\
		\midrule
		Town01   & $800$  & $309$  & $798$    & $1572$  & $100$ \\
		Town02   & $800$  & $577$  & $898$    & $1983$  & $100$ \\
		Town03   & $800$  & $581$  & $1574$   & $3471$  & $100$ \\
		Town04   & $792$  & $555$  & $3167$   & $5978$  & $99$  \\
		Town05   & $800$  & $695$  & $1727$   & $3855$  & $100$ \\
		Town06   & $800$  & $229$  & $445$    & $2495$  & $100$ \\
		Town07   & $800$  & $251$  & $758$    & $1967$  & $100$ \\
		Town10   & $792$  & $823$  & $1648$   & $2998$  & $99$  \\
		Total    & $6384$ & $4020$ & $11015$  & $24319$ & $798$ \\
		\bottomrule
	\end{tabular}
	\caption{Overview of CARLA3D dataset. \textit{Frames} represents the number of point cloud frames sampled in the
		scenario; \textit{Easy}, \textit{Moderate}, and \textit{Hard} represent the quantities of objects with different difficult levels in the scenario, respectively. \textit{Times} refers to the number of sampling.}	\label{tab:CARLA3D_dataset}
\end{table}
\begin{table}[tbp]
	\centering
	\resizebox{\columnwidth}{!}{
		\begin{tabular}{cccc}
			\toprule
			Dataset                                        & Size(Train/Test) & LiDAR Beams                & Points Per Frame \\
			\midrule
			CARLA3D                                        & 3990 / 2394      & $1\times64$                & $286.2K$         \\
			KITTI\cite{geigerAreWeReady2012}               & 3712 / 3769      & $1\times64$                & $118.7K$         \\
			Lyft\cite{kestenLyftLevelPerception2019}       & 12017 / 2891     & $1\times40\textit{ or }64$ & $72.3K$          \\
			TinySUScape\cite{dingJstJointSelftraining2022} & 2579 / 965       & $1\times128$               & $230.4K$         \\
			\bottomrule
		\end{tabular}
	}
	\caption{A summary of datasets. The \textit{Size(Train/Test)} refers to the number of samples used in training and testing.}
	\label{tab:datasets}
\end{table}

\begin{table}[!t]
	\centering
	\begin{tabular}{@{}ccc|c|cc|c@{}}
		\toprule
		AH           & Aug2         & MT           & NLL & FL-NA        & OL-NA        & $mAP_{3D}$     \\ \midrule
		             &              &              &     &              &              & 15.44          \\ \midrule
		$\checkmark$ &              &              &     &              &              & 34.63          \\
		$\checkmark$ & $\checkmark$ &              &     &              &              & 43.51          \\
		$\checkmark$ & $\checkmark$ &              & CF  &              &              & 43.83          \\\midrule
		$\checkmark$ & $\checkmark$ & $\checkmark$ &     &              &              & 45.67          \\
		$\checkmark$ & $\checkmark$ & $\checkmark$ & CF  &              &              & 45.91          \\\midrule
		$\checkmark$ & $\checkmark$ & $\checkmark$ & CF  & $\checkmark$ &              & 46.47          \\
		$\checkmark$ & $\checkmark$ & $\checkmark$ & CF  &              & $\checkmark$ & 48.67          \\
		$\checkmark$ & $\checkmark$ & $\checkmark$ & BF  & $\checkmark$ & $\checkmark$ & 49.37          \\

		$\checkmark$ & $\checkmark$ & $\checkmark$ & CF  & $\checkmark$ & $\checkmark$ & \textbf{50.56} \\


		\bottomrule
	\end{tabular}

	\caption{Ablation study results on CARLA3D $\to$ Lyft. \textit{AH}: anchor head scheme proposed in Sec \ref{sec:anchor_head}; \textit{Aug2}: second-stage augmentation in Sec \ref{sec:roi_augmentation}; \textit{MT}: mean teacher based domain adaptation; \textit{NLL}: usage of NLL loss for aleatoric uncertainty; \textit{CF} and \textit{BF} refer to corner-format and box-format encoding respectively in Sec \ref{sec:detection_au}; \textit{FL-NA} and \textit{OL-NA}: frame-level and object-level noise-aware sampling strategies respectively in Sec \ref{sec:noise-aware sampling}. The $mAP_{3D}$ metric is obtained by averaging over the three difficulty levels.}
	\label{table:ablation_study_res}

\end{table}

The results obtained using different UDA methods are summarized in Table \ref{table:main_results}. Our CTS method surpasses all others in sim-to-real detection tasks. Specifically, compared to the \textit{source only} method, our approach improves $AP_{BEV}$ by approximately $15\% - 35\%$ and $AP_{3D}$ by around $25\% - 50\%$. However, due to the significant domain shift between the simulator and reality, our CTS method still exhibits a noticeable gap compared to the supervised \textit{Oracle}.
In contrast, the SN method, which generally performs well in various real-world domains, struggles in sim-to-real cross-domain tasks, experiencing performance degradation, such as in the CARLA3D $\to$ TinySUSCape scenario.

\begin{figure}
	\centering
	\includegraphics[width=0.7\columnwidth]{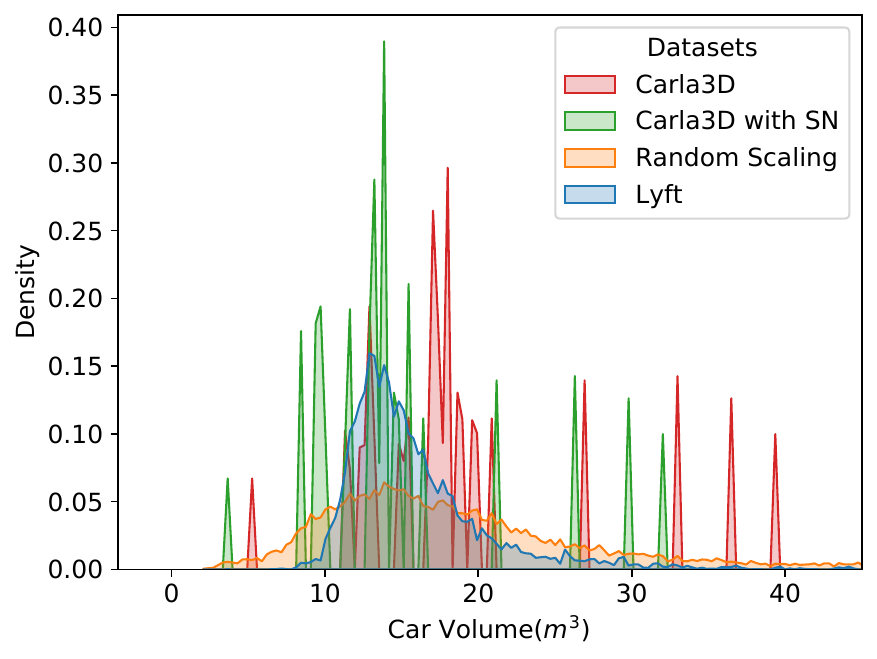}
	\caption{An illustration of the car sizes distribution of Lyft\cite{kestenLyftLevelPerception2019}, and CARLA3D datasets with different processing methods, \textit{i.e.}, SN\cite{wangTrainGermanyTest2020} and Random Scaling.}\label{fig:stat_size_sn}
\end{figure}

\subsection{Ablation Study}

To further demonstrate the effectiveness of the individual components in our proposed method, we conducted extensive ablation experiments on the CARLA3D $\to$ Lyft task.

\subsubsection{Benefits of Anchor Head}
Incorporating the anchor head (AH) into the second-stage detector effectively reduces regression complexity while enhancing cross-domain robustness. As described in Table \ref{table:ablation_study_res}, compared to the original setup, the AH scheme yields over $19\%$ improvement, highlighting its effectiveness in cross-domain tasks even with a simple anchor size replacement.

\subsubsection{Benefits of RRS and Second-stage Augmentation}
Compared to SN's approach \cite{wangTrainGermanyTest2020}, our RoI Random Scaling (RRS) method effectively encourages the sizes of processed objects to resemble an unimodal distribution similar to real-world data, rather than solely aligning with statistical volumes that still exhibit multi-modal, as illustrated in Figure \ref{fig:stat_size_sn}.
Furthermore, integrating RRS into our second-stage augmentation (Aug2) resulted in a performance improvement of approximately $9\%$, as demonstrated in Table \ref{table:ablation_study_res}. These augmentation techniques enhance data diversity at the object level, enabling the model to learn diverse information.

\subsubsection{Benefits of Corner-Format AU}
\begin{figure}[]
	\vspace{-9pt}  
	\centering
	\subfloat[]{
		\centering
		\includegraphics[width=0.47\columnwidth,clip]{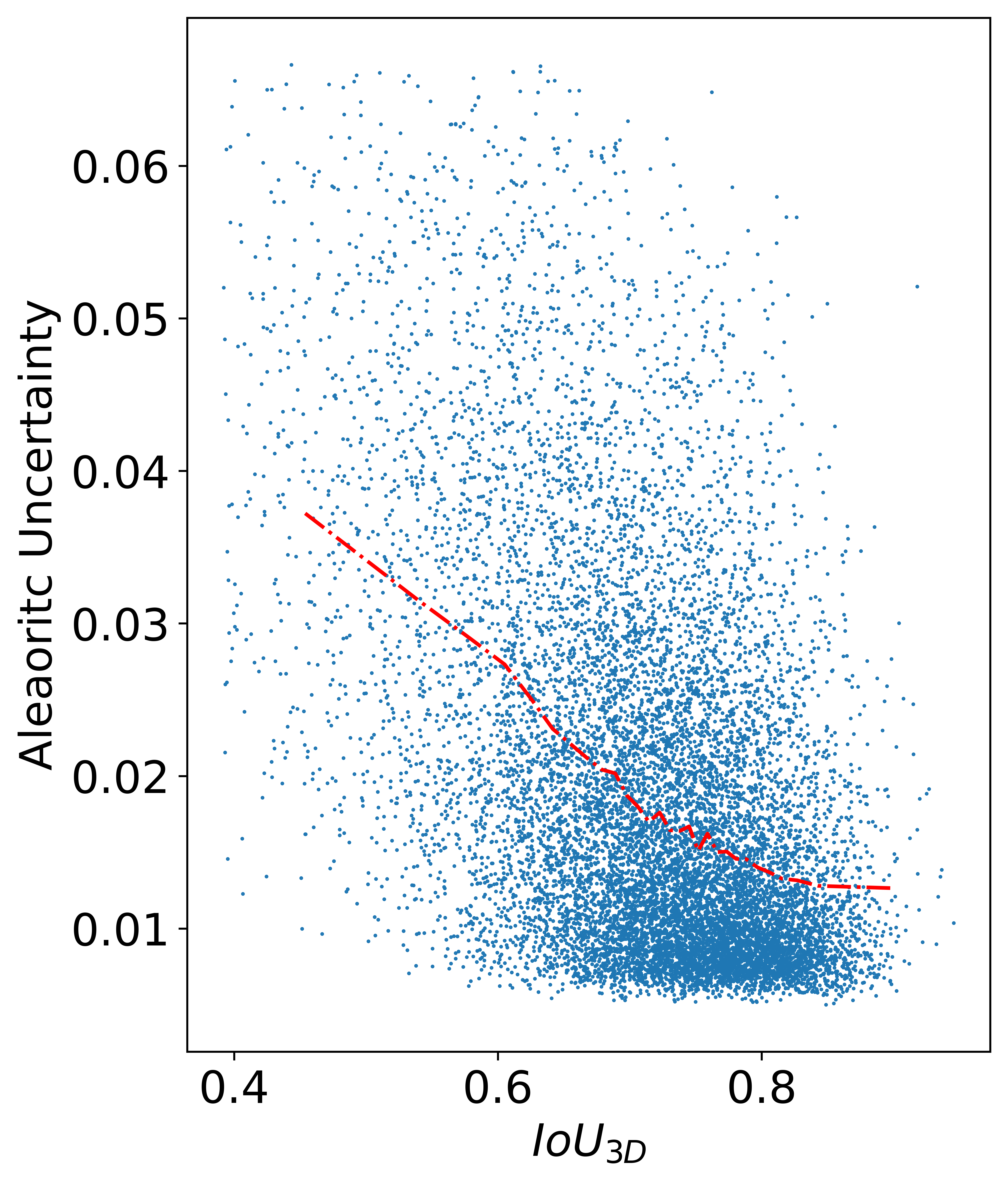}
	}
	\hfill
	\subfloat[]{
		\centering
		\includegraphics[width=0.47\columnwidth,clip]{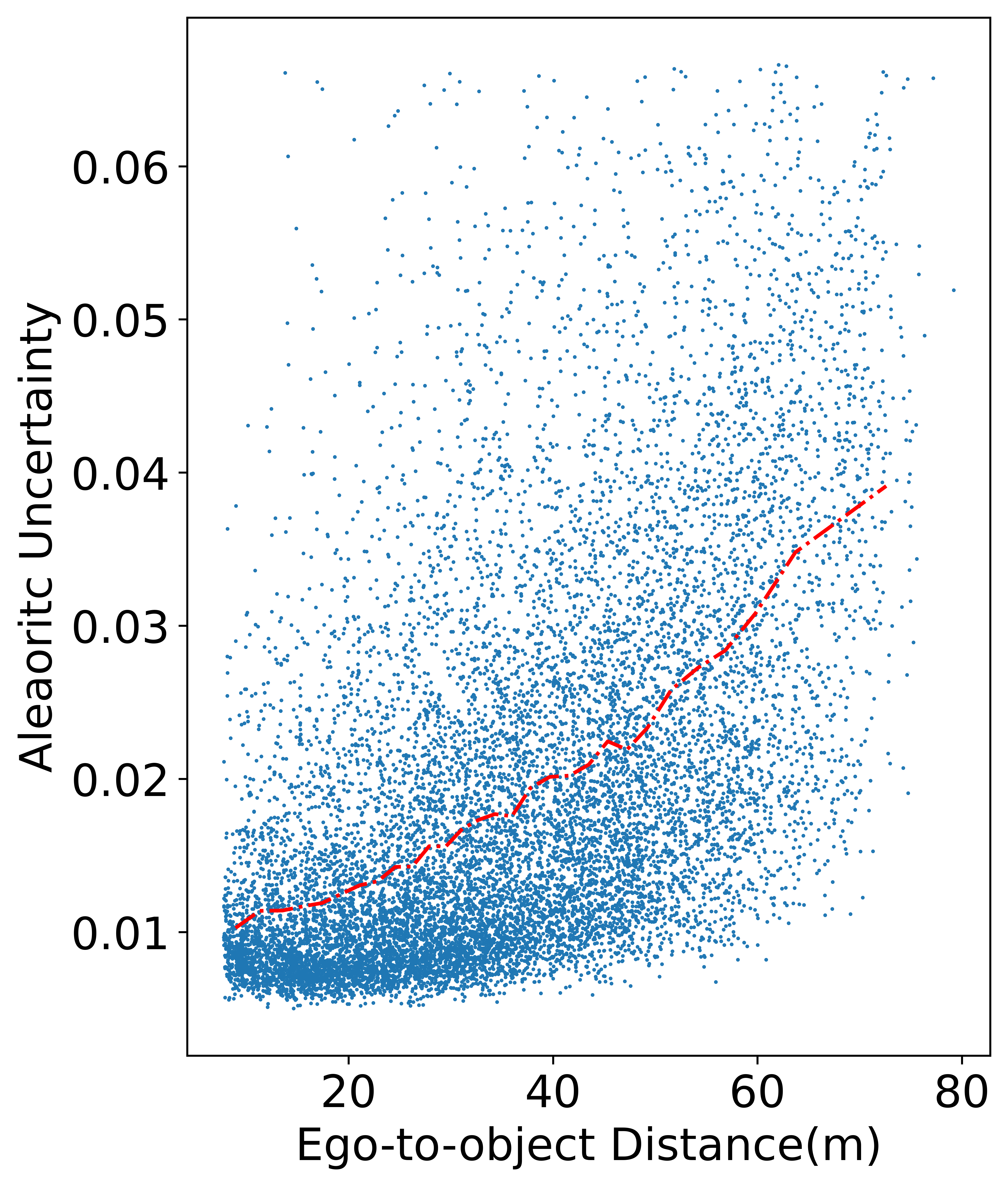}
	}
	\vspace{-9pt}  

	\caption{
		An illustration of the correlation between AU value and IoU/ego-to-object distance for the target dataset. \textit{Blue} points denote the AU values of detected objects; the \textit{red} line represents the means of the AU values.}
	\label{fig:iou_uncer}
\end{figure}

\begin{figure}[]
	\centering
	\subfloat[Easy, AU$=0.007$]{%
		\includegraphics[width=0.35\linewidth]{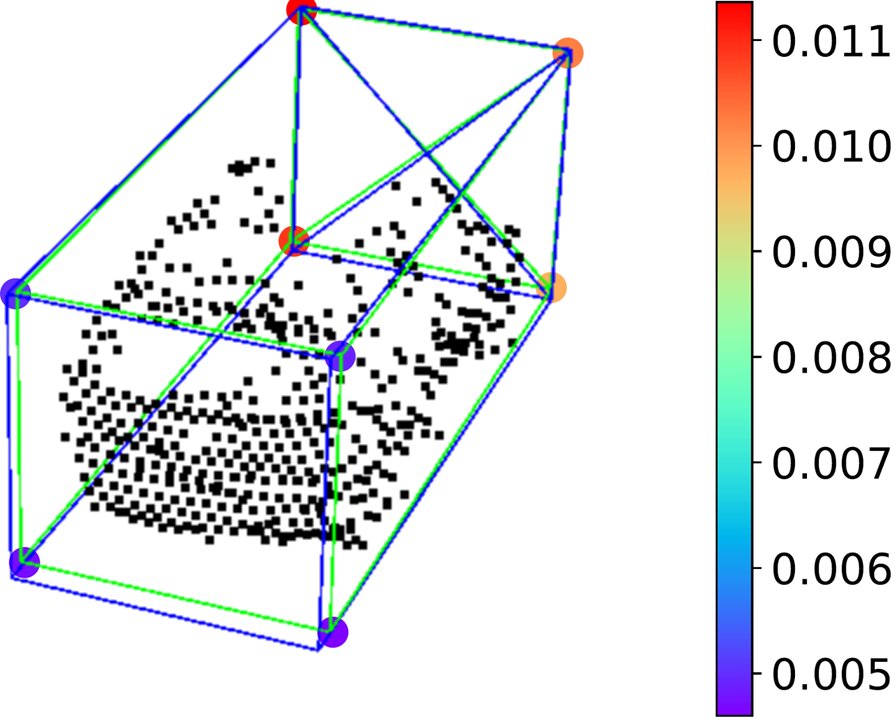}}
	\hspace{12pt}
	\subfloat[Moderate, AU$=0.018$]{%
		\includegraphics[width=0.35\linewidth]{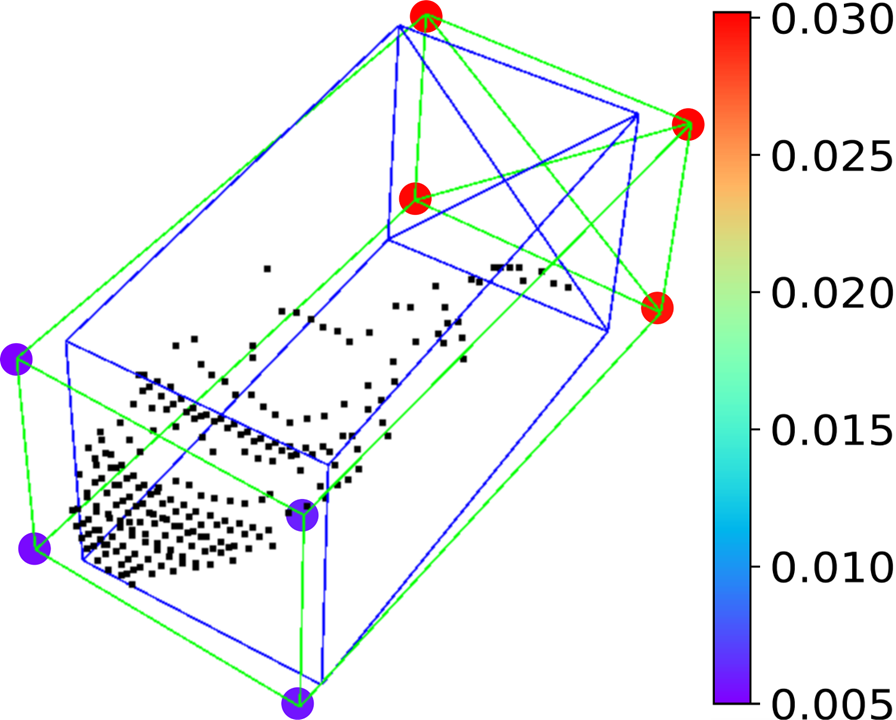}}
	\\
	\subfloat[Moderate, AU$=0.016$]{%
		\includegraphics[width=0.35\linewidth]{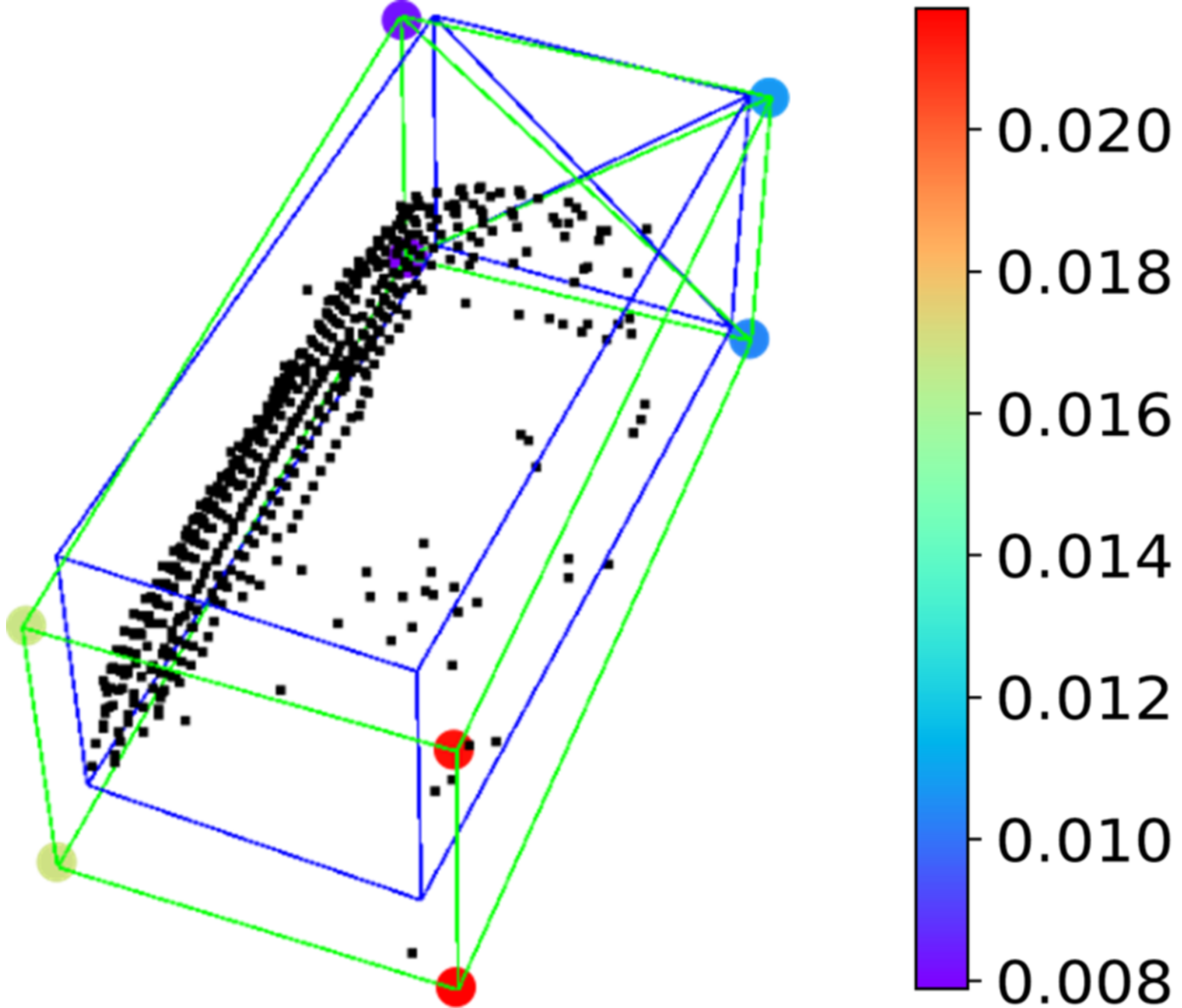}}
	\hspace{12pt}
	\subfloat[Hard, AU$=0.060$]{%
		\includegraphics[width=0.35\linewidth]{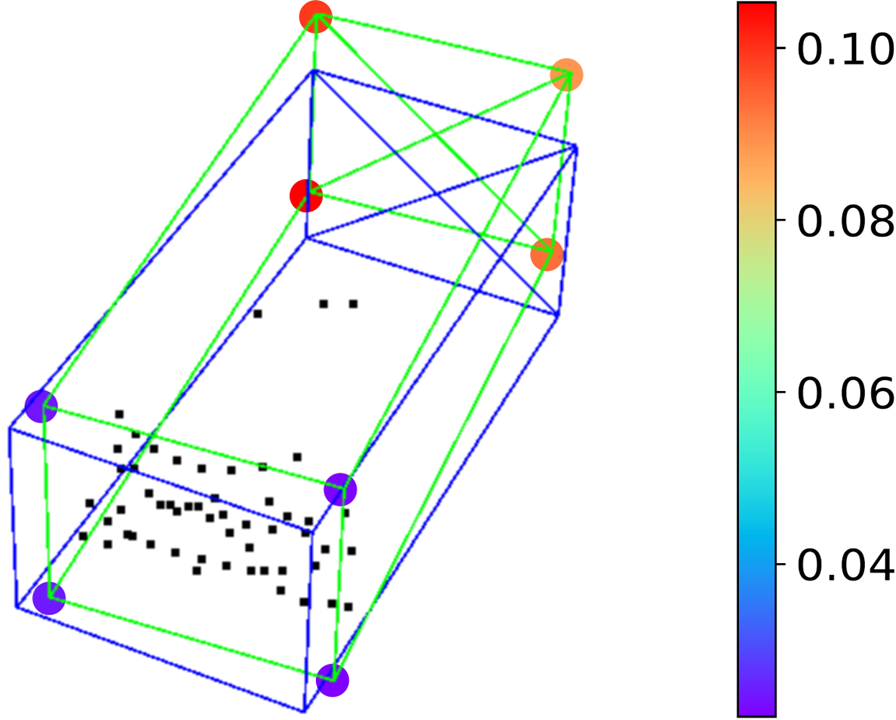}}
	\caption{Examples of different levels of difficulties in 3D boxes. The \textit{blue} boxes represent the ground truth; the \textit{green} boxes represent the predicted results. The points in different colors at the box corners represent the 8 AU value components, whose mean is the final AU value of the entire object.}
	\label{fig:cases}
\end{figure}

In contrast to BF, CF encoding uniformly distributes the localization uncertainty of the object across each corner component without requiring additional operations. Table \ref{table:ablation_study_res} demonstrates that using BF and CF representations for noise-aware sampling improves performance by $3.7\%$ and $4.9\%$, respectively. This suggests that CF is more effective in identifying reliable pseudo-labels.
Employing the CF encoding scheme, we investigate the aleatoric uncertainties (AUs) associated with predicted objects, considering their Intersection over Union (IoU) with ground truths and their ego-to-object distance, as depicted in Figure \ref{fig:iou_uncer}. Our observations reveal a decrease in AU values with increasing IoU, while they increase with greater ego-to-object distance. Furthermore, Figure \ref{fig:cases} showcases examples where sparse and corrupted point clouds lead to elevated AU. These findings underscore the efficacy of predicted AUs in evaluating pseudo-label noise and their utility as a reliability metric for pseudo-labels.

\subsubsection{Benefits of Noise Awareness in Mean Teacher}
As mentioned in Sec \ref{sec:noise-aware sampling}, two diverse noise-aware sampling strategies are used to minimize the adverse impacts of noisy pseudo-labels generated during mean teacher domain adaptation. with both the \textit{frame-level noise-aware} (FL-NA) and \textit{object-level noise-aware} (OL-NA) strategies, performance improves by $4.65\%$.

Additionally, utilization of NLL loss function solely has been shown to bring improvement \cite{fengLeveragingHeteroscedasticAleatoric2019}. Table \ref{table:ablation_study_res} also indicates a minor increase from 43.51\% to 43.81\% in source-only training with NLL. However, While adding NLL loss and extra uncertainty layers yields only a $0.3\%$ improvement, employing both FL-NA and OL-NA results in an extra significant improvement of $4.3\%$. This demonstrates that the main performance gain arises from noise-aware sampling strategies rather than just loss function replacement.

\subsection{Limitations}
Although our proposed model shows enhanced adaptation performance within the target domain via multiple schemes, sim-to-real UDA still lags behind real-to-real methods due to limitations inherent in simulators. The restricted vehicle assets in simulators like CARLA fail to represent the diverse range of real-world vehicles. Additionally, simulators struggle to replicate complex real-world scenarios, including dynamic traffic patterns and diverse urban landscapes (e.g., different weather conditions), thus limiting their effectiveness in providing realistic training data for domain adaptation.

\section{Conclusion}\label{sec:conclusion}

This paper has introduced a CTS framework for unsupervised domain adaptation (UDA) in 3D object detection, bridging the gap between simulation and real-world domains. The proposed techniques, including RoI random scaling and augmentation, along with the fixed-size anchor head, enhance the diversity of simulation data and address object size discrepancies across domains, thereby improving the quality of pseudo-labels. Additionally, the proposed aleatoric uncertainty (AU) estimation, based on a uniform corner-format representation of bounding boxes, facilitates the integration of pseudo-label noise awareness into the mean teacher domain adaptation process, leading to high-quality pseudo-label sampling. Experimental results on the CARLA, KITTI, Lyft, and TinySUScape datasets demonstrate substantial improvements over existing methods in various sim-to-real UDA tasks, with 5\%-17\% gains in $AP_{3D}$ and 2\%-10\% gains in $AP_{BEV}$. Future work will focus on extending this approach to cover both sim-to-real and real-to-real UDA scenarios, as well as incorporating additional categories (e.g., bicycles, pedestrians) in domain adaptation.

\bibliographystyle{IEEEtran}
\bibliography{IEEEabrv, reference}
\end{document}